\documentclass{article}
\usepackage{times}
\usepackage{proceedings}
\usepackage{itmath}
\usepackage{latexsym,amssymb,stmaryrd}
\usepackage{epsfig}        % Include eps figure
\def\set#1{\{#1\}}              % \set{1, 2, 3} produces {1, 2, 3}
\def\tuple#1{\langle#1\rangle}  % \tuple{1, 2, 3} produces <1, 2, 3>
\def\And{\wedge}                % produces ^
\def\AND{\bigwedge}             % produces big ^
\def\Or{\vee}                   % produces v
\def\Nat{I\!N}                  % natural numbers
\def\qsat2{\mbox{{\sc qsat}}_2}
\def\PEC{{\mathcal P}{\mathcal E}}
\def\SEC{{\mathcal S}{\mathcal E}}
\def\BEC{{\mathcal B}{\mathcal E}}
\def\HC{{\mathcal H}}
\def\XC{{\mathcal X}}
\def\AC{{\mathcal A}}

\newtheorem{definition}{Definition}
\newtheorem{theorem}{Theorem}
\newtheorem{corollary}{Corollary}
\newtheorem{problem}{Problem}

\newenvironment{proof}{\par\vspace*{0.5ex}\noindent\rm {\em Proof.} }%
         {{\hfill\tiny$\Box$}\par\vspace*{1ex}}
\begin{document}
%\begin{frontmatter}
\title{On Concise Encodings of Preferred Extensions}
\author{Paul E. Dunne \\
Department of Computer Science \\
University of Liverpool \\
Liverpool L69 7ZF \\
United Kingdom\\
{\tt ped@csc.liv.ac.uk}
}

\maketitle

\begin{abstract}
Argument Systems provide a rich abstraction within which divers concepts of
reasoning, acceptability and defeasibility of arguments, etc., may be studied
using a unified framework. Much work has focused on
the so-called {\em preferred} extensions of such systems, which define the maximal
(with respect to $\subseteq$) collectively defensible subsets of arguments within
a given system of arguments and attack relationship. In this article we address
problems related to the following issue. Identification and enumeration of preferred extensions
of an argument system is (under the usual complexity theoretic assumptions) computationally
demanding: there may be exponentially many and deciding if a given subset $S$ of
${\mathcal X}$ {\em does} define a preferred set is {\sc co-np}--complete. 
For a domain which is questioned `frequently' it may be acceptable to invest this
computational effort once, but having done so it is desirable to encapsulate these data
in a form which is compact and allows, for example, questions concerning the acceptability
of specific arguments to be dealt with efficiently. In this article we consider two `plausible'
approaches to reducing the complexity of deciding if $S$ is a preferred extension
of a system ${\mathcal H}$ both of which assume some initial potentially `expensive'
precomputation, invested to reduce time needed in subsequent queries to the system. The first approach
examines `reasonable encoding' approaches; the second is to determine the subset of
all defensible arguments providing these as additional data when attempting to
decide if $S$ is a preferred extension. It is shown that if certain properties
are required of the encoding scheme, then the former approach is feasible only if
$\mbox{{\sc np}}=\mbox{{\sc co-np}}$. In the latter case, we show that, 
even if provided with information regarding which arguments are credulously accepted, 
the question of whether a subset of arguments defines a preferred extension remains {\sc co-np}--complete.
\end{abstract}

{\bf Keywords:} Argument Systems, Preferred Extension, Computational Complexity
%\end{frontmatter}
%\newpage
\section{Introduction}
Since they were introduced by Dung~\cite{dung:1995}, Argument Systems have provided
a fruitful mechanism for studying reasoning in defeasible contexts. They have proved 
useful both to theorists who can use them as an abstract framework for the study and 
comparison of non-monotonic logics, 
e.g.~\cite{bondar-etal:1997}, 
and for those who wish to explore more concrete contexts where defeasibility is central. 
In the study of reasoning in law, for example, they have been used to examine the 
resolution of conflicting norms, e.g. \cite{prakken:1997}, 
especially where this is studied through the mechanism of a dispute between two 
parties, e.g. \cite{JakobVerm99}. 
The basic definition below is derived from that given in \cite{dung:1995}.

\begin{definition}\label{defn:basicform}
An {\em argument system} is a pair $\mathcal{H} = \tuple{\mathcal{X},\mathcal{A}}$, in which
${\mathcal{X}}$
is a set of {\em arguments} and ${\mathcal{A}} \subset {\mathcal{X}} \times {\mathcal {X}}$
is the {\em attack relationship} for $\mathcal{H}$.
Unless otherwise stated, ${\mathcal X}$ is assumed to be {\em finite}, and ${\mathcal A}$
comprises a set of ordered pairs of {\em distinct} arguments.
A pair $\tuple{x,y}\in\mathcal{A}$ is referred to as `$x$ {\em is attacked by} $y$' or `$y$ {\em attacks}
(or {\em is an attacker of}) $x$'.

For $R$, $S$ subsets of arguments in the system $\mathcal{H}(\tuple{\mathcal{X},\mathcal{A}})$,
we say
that
\begin{enumerate}
\item[a)]
$s\in S$ is {\em attacked} by $R$ if there is some $r\in R$ such that $\tuple{r,s}\in\mathcal{A}$.
\item[b)]
$x\in \mathcal{X}$ is {\em acceptable with respect to} $S$ if for every $y\in \mathcal{X}$ that attacks $x$
there is some $z\in S$ that attacks $y$.
\item[c)]
$S$ is {\em conflict-free} if no argument in $S$ is attacked by any other argument in $S$.
\item[d)]
A conflict-free set $S$ is {\em admissible} if every argument in $S$ is acceptable with respect
to $S$.
\item[e)]
$S$ is a {\em preferred extension} if it is a maximal (with respect to $\subseteq$) admissible
set.
\item[f)]
$S$ is a {\em stable extension} if $S$ is conflict free and every argument $y\not\in S$ is
attacked by $S$.
\item[g)]
${\mathcal H}$ is {\em coherent} if every preferred extension in ${\mathcal H}$ is also a stable
extension.
\end{enumerate}
An argument $x$ is {\em credulously accepted} is there is {\em some} preferred extension containing
it; $x$ is {\em sceptically accepted} if it is a member of {\em every} preferred extension.

The notation ${\mathcal P}{\mathcal E} ({\mathcal H})$ is used to describe the set of
all subsets of ${\mathcal X}$ which are preferred extensions of ${\mathcal H}$. Similarly,
$\SEC (\HC)$ denotes the set of all {\em stable} extensions of $\HC$ and $\BEC(\HC)$ refers
to an arbitrary one of these sets.
We use $n$ to denote $|{\mathcal X}|$.
\end{definition}

The preferred extensions of an
Argument System can be taken as being the consistent positions that can be adopted within the Argument System. 
Any argument that appears in all preferred extensions will be acceptable in every consistent position, 
and any argument that appears in no preferred extension cannot be held in any consistent position. 
This means that the notion can be related to varieties of semantics for non-monotonic reasoning: 
credulously acceptable arguments will be those that appear in at least one preferred extension, and 
sceptically acceptable arguments will be those which appear in all preferred extensions. 
In the context of legal reasoning, the notion allows us to distinguish those arguments which must be 
accepted, those which can be defended, and those which are indefensible.

To avoid repetition we will subsequently refer to the following decision problems:
\newline
\newline
{\sc pref-ext} (Preferred Extension)
\newline
{\bf Instance}: An argument system $\HC({\mathcal X},{\mathcal A})$ and $S\subseteq{\mathcal X}$.
\newline
{\bf Question}: Is $S\in\PEC(\HC)$?
\newline
{\sc stab-ext} (Stable Extension)
\newline
{\bf Instance}: An argument system $\HC({\mathcal X},{\mathcal A})$ and $S\subseteq{\mathcal X}$.
\newline
{\bf Question}: Is $S$ a subset of some  $T\in\SEC(\HC)$, i.e. can $S$ be expanded to a stable extension?
\newline
{\sc pref-ext-inf} (Preferred Extension given Information)
\newline
{\bf Instance}: An argument system $\HC({\mathcal X},{\mathcal A})$, $S\subseteq{\mathcal X}$, and 
$\alpha=\tuple{a_1,a_2,\ldots,a_n}\in\tuple{\bot,\top}^n$ an $n$-tuple of Boolean values
such that $a_i=\top$ if and only if the argument $x_i$ is credulously accepted in $\HC$.
\newline
{\bf Question}: Is $S\in\PEC(\HC)$?
\newline
{\sc stab-ext-inf} (Stable Extension given Information)
\newline
{\bf Instance}: An argument system $\HC({\mathcal X},{\mathcal A})$, $S\subseteq{\mathcal X}$, and 
$\alpha=\tuple{a_1,a_2,\ldots,a_n}\in\tuple{\bot,\top}^n$ an $n$-tuple of Boolean values
such that $a_i=\top$ if and only if the argument $x_i$ is credulously accepted in $\HC$.
\newline
{\bf Question}: Is $S$ a subset of some  $T\in\SEC(\HC)$, i.e. can $S$ be expanded to a stable extension?

Before proceeding it may be useful to consider in more detail the concepts of `preferred' versus
`stable' extensions of an argument system. Both \cite{dung:1995} and \cite{bondar-etal:1997}
offer a view of preferred extensions as providing a more general construct than
stable extensions: thus any stable extension is also preferred but the converse is not
always true. A significant difference between the two models is that whereas some argument
systems may have no stable extension, it is always the case that a preferred extension
extension exists since the empty set is always admissible. This difference raises a number
of general questions that are discussed in some detail in \cite{dung:1995}. In particular,
\begin{enumerate}
\item
Are there `natural' or `meaningful' argument systems with no stable extensions?
\item
Are there `natural' systems whose set of stable extensions form a strict (non-empty) subset of the
set of preferred extensions?
\end{enumerate}
Through consideration of a particular $n$-player game, \cite[p.336]{dung:1995}, argues
that studies of Lucas and Shubik\cite{Shubik:1985} suggest `{\em stable semantics do not
capture the intuitive semantics of every meaningful argumentation system}'. As a further
example, using a variant of the {\em Stable Marriage Problem}, \cite[p.338]{dung:1995} exhibits
a concrete `natural' system which has no stable extension. While both examples suggest
a positive answer to the first of the two questions raised, neither treats the second question.
Instead, \cite[Defn.~31,~p.~332]{dung:1995} introduces the concept of {\em coherence}
to describe systems for which every preferred extensions is also stable, discussing
forms of argumentation system whose instantiation guarantees coherence. Recent work of
Dunne and Bench-Capon\cite{dunnebc:2001c}, however, indicates that even when restricted to
the context of finitely presented argument systems, deciding if a given system 
$\HC(\XC,\AC)$ is coherent is `likely' to be extremely hard. One consequence of the 
proof employed in \cite{dunnebc:2001c}, is that it, naturally, gives rise to an infinite
class of argument systems having a non-empty set of stable extensions but which are nonetheless
incoherent: i.e. there is a supportable case that the answer to the second question raised
is also positive.

A major difficulty that is encountered within these formalisms is the computational intractability 
of several decision problems that arise: 
results of Dimopoulos and Torres~\cite{DimopTorres:1996} indicate that deciding if $p$
is credulously accepted in ${\mathcal H}$ is {\sc np}--complete and that {\sc pref-ext}
is {\sc co-np}--complete (even when $S$ is the empty set). Similarly
deciding if $\HC$ has any stable extension (i.e. the decision problem $\SEC(\HC)\not=\emptyset?$)
is $\mbox{{\sc np}}$-complete (notice that this is the special case of {\sc stab-ext} when $S=\emptyset$).  
Dunne and Bench-Capon~\cite{dunnebc:2001c} proved
that deciding coherence is $\Pi_{2}^{(p)}$--complete, deducing the same complexity classification
for sceptical acceptance as a consequence. Related work, described in \cite{dunnebc:2001}, has shown
that the sound and complete reasoning method for credulous argumentation introduced by
Vreeswijk and Prakken \cite{Vrees-Prakk:2000} in which reasoning proceeds via
a dialogue game requires an exponential number of moves to resolve some disputes. Examining differing
concepts of `acceptance' in various non-monotonic Logics, \cite{dimop-etal:1999,dimop-etal:2000},
indicate that decision problems predicated on preferred extensions are complete for divers
levels of the polynomial--time hierarchy ranging from $\Sigma_{1}^{p}$ (i.e. {\sc np}) in the
case of credulous reasoning in the Logic Programming formalism ({\sc lp}) to $\Pi_{4}^{p}$
in the case of sceptical reasoning within Autoepistemic Logic ({\sc ael}).

The concern of this paper is to consider some further contrasts between computational properties
of preferred and stable extension sets, arising from the following scenario.
Suppose one is given a specific argument system ${\mathcal H}({\mathcal X},{\mathcal A})$ which
describes a `frequently' used application, for example one incorporating complex legal data
in which the justification for various different positions may have to be assessed. Rather than
deciding acceptability and support for an argument on each new query to ${\mathcal H}$, we may be
prepared to invest some computational effort {\em once} in the hope that
the information elicited as a result may help in reducing the time taken for subsequent queries.
For example, one could compute all sets in $\BEC(\HC)$  and then have queries
on ${\mathcal H}$ performed with respect to a {\em representation} (encoding) of this set of subsets. 
There are, of course, several properties which such a representation should, ideally, satisfy. 
Suppose $\eta({\mathcal H})$
is some encoding of ${\mathcal B}{\mathcal E}({\mathcal H})$. We define, informally,
two such properties that are the main focus of this paper:
\begin{enumerate}
\item[R1.]
$\eta$ is {\em terse}, i.e. $|\eta({\mathcal H})|$ -- the number of {\em bits} needed --
is polynomially bounded in $n$.
\item[R2.]
$\eta$ is {\em extension tractable}, i.e. given any $S\subseteq{\mathcal X}$, 
the question $S\in{\mathcal B}{\mathcal E}$ can be decided from
$\eta({\mathcal H})$ in time polynomial in $|\eta({\mathcal H})|$.
\end{enumerate}

One obvious representation scheme is simply to use a table $T_{{\mathcal H}}$ of
$|{\mathcal B}{\mathcal E} ({\mathcal H})|$ rows, each row being $n$ bits in length, so
that if $\tuple{S_1,\ldots,S_k}$ is an ordering of $\BEC(\HC)$, then $T_{i,j}=1$ if and only if $x_j\in S_i$.
While this representation meets the criterion specified by R2, it will fail to satisfy R1 in those
cases where $|{\mathcal B}{\mathcal E}({\mathcal H})|$ is superpolynomial in $n$. We note
that systems may be defined where this number is $\Omega (3^{n/3})$.
Alternatively, the system ${\mathcal H}$ itself is a representation of
${\mathcal B}{\mathcal E} ({\mathcal H})$: while satisfying R1 it is, however, unlikely to satisfy
R2 in the case $\BEC=\PEC$ (assuming $\mbox{{\sc np}}\not=\mbox{{\sc co-np}}$).

Given such examples, a natural question to raise is whether these extremes are inherent, or
are there representation formalisms that are terse {\em and} extension tractable -- a property we subsequently
refer to as {\em concise}.
In the next Section we formalise these concepts and, in Section 3, prove some
basic results concerning them.
In particular it is shown that in the case of $\PEC(\HC)$ `effective' concise encodings are not possible, 
in general,
unless $\mbox{{\sc np}}=\mbox{{\sc co-np}}$. In contrast, concise encoding schemes for $\SEC(\HC)$
are easy to construct. We note that this provides another
example of a property which is considerably `easier' under stable semantics than under
preferred semantics for argument systems, cf. \cite{dimop-etal:1999,dimop-etal:2000}.
A further indication of the computational difficulties arising in considering preferred
extensions is given in the concluding result of Section 3 where the problem {\sc pref-ext-inf}
is shown to be {\sc co-np}--complete: thus, even if the defensibility status for every argument
in ${\mathcal X}$ is supplied as part of an instance, the problem of deciding whether a given subset $S$
is a preferred extension does not become any easier, i.e. remains {\sc co-np}--complete. Discussion and
conclusions occupy Section 4.

\section{Definitions}
In the remainder of this paper the following notational conventions are used.

${\mathcal X}_n$ is a set of $n$ arguments $\set{x_1,x_2,\ldots,x_n}$.

For an $n$ element set $X_n$, $\wp (X_n)$ is the set of all subsets of $X_n$.

The following definition formalises our abstract concept of {\em encoding scheme} for
the set $\BEC$ in an argument system $\HC$.
\begin{definition}\label{defn:encode}
A {\em $\BEC$ encoding scheme} is a pair $\tuple{\eta,P}$ where
\[
\eta : \set{\HC({\mathcal X}_n,{\mathcal A})\mbox{ : $\HC$ is an argument system }}
\rightarrow\set{0,1}^*
\]
is a mapping from argument systems to finite binary words, and $P$ is a {\em deterministic}
Turing machine program, that takes as input a pair $\tuple{\eta(\HC),S}$ in which $S\subseteq{\mathcal X}_n$,
accepting if and only if $S\in\BEC(\HC)$.
\end{definition}

Before proceeding there are several points that should be noted. First we observe
that it is not insisted that schemes be {\em uniform}, i.e. it is {\em not} required
that there is an algorithm which given $\HC$ computes $\eta({\HC})$. Of course, in
`practical' schemes, one would wish to have some mechanism for automating this translation.
In order to capture some sense of `practical' scheme, we introduce the notion of {\em verifiability}
\begin{definition}
A {\em verifiable} $\BEC$ {\em encoding scheme} is a triple $\tuple{\eta,P,Q}$ in which
$\tuple{\eta,P}$ is a $\BEC$ encoding scheme and $Q$ a (non-deterministic) Turing Machine program,
that is given $w\in\set{0,1}^*$ and $\HC({\mathcal X},{\mathcal A})$ as input,
accepting if $w=\eta(\HC)$.
\end{definition}
Secondly, the definition provides a foundation for introducing more `sophisticated' schemes other
than binary words. 

With this abstract idea of encoding scheme we can formalise the notions of {\em terse} and {\em extension
tractable} outlined earlier.
\begin{definition}
Let $\tuple{\eta,P,Q}$ be a verifiable $\BEC$ encoding scheme.
We say $\tuple{\eta,P}$ is {\em terse} if there is a constant $k$, such that for all 
$\HC({\mathcal X}_n,{\mathcal A})$, $|\eta({\mathcal H})|\leq n^k$; it is {\em extension tractable} if there 
is a constant $k$ such that: given $\eta({\mathcal H})$ and $S\in\wp(X_n)$ as input,
$P$ decides if $S\in \BEC(\HC)$ taking at most $(n+|\eta(\HC)|)^k$ steps. Finally,
$\tuple{\eta,P,Q}$ is {\em effective} if there is a constant $k$ for which $Q$ has an accepting
computation of $w=\eta(\HC)$ in non-deterministic time bounded by $(n+|w|)^k$.
A $\BEC$ encoding scheme $\tuple{\eta,P}$ is {\em concise} if it is both terse and extension tractable.
A {\em verifiable} $\BEC$ encoding scheme $\tuple{\eta,P,Q}$ is {\em usefully concise} if it is
terse, extension tractable, and effective.
\end{definition}

In `practical' terms {\em usefully concise} verifiable $\BEC$ encoding schemes, define the `ideal'
representation form: if $\tuple{\eta,P,Q}$ is usefully concise then one can describe $\BEC(\HC)$
in its entirety using only a `small' amount of space -- since $\eta$ is {\em terse}; one can determine 
efficiently (in terms of $|\eta(\HC)|$ and $|S|$) if $S\in\BEC(\HC)$ -- since the scheme is
{\em extension tractable}; and, finally, one may test if an arbitrary $w$ does, indeed, describe
the encoding $\eta(\HC)$ for a given $\HC$.

We now present some examples of encoding approaches.

\subsubsection{Tabular Representation}
Given ${\HC}$, $tab ({\mathcal H})$ is the $n|\BEC(\HC)|$-bit word
in which bit $t_{(i-1)n+j}=1$ if and only if $x_j\in S_i$ where $1\leq j\leq n$ and
$\tuple{S_1,S_2,\ldots,S_r}$ is an ordering of $\BEC(\HC)$. If the algorithm $P$ in $\tuple{tab,P}$
is chosen to be an appropriate table look up method, then $\tuple{tab,P}$ is extension tractable.
It is not, however, terse.

\subsubsection{Representations via Propositional Logic Functions}
Given any $\HC({\mathcal X}_n,{\mathcal A})$ there is a unique propositional
logic function, $f_{\HC} (X_n)$ definable from $\BEC(\HC)$ as follows. For any $S\in\wp({\mathcal X}_n)$
let the instantiation, $\alpha_S$ of the propositional variables $X_n$ be
$x_i=\top$ if $x_i\in S$ and $x_i=\bot$ if $x_i\not\in S$. The function $f_{\HC} (X_n)$ takes the
value $\top$ on exactly those instantiations $\alpha_S$ for which $S\in\BEC(\HC)$.

Given this approach, any representation formalism for arbitrary $n$-argument propositional logic
functions serves as a basis for a $\BEC$ encoding scheme, e.g. truth-tables,
propositional formulae over a finite complete basis, etc.

\section{Properties of Usefully Concise Encoding Schemes}

We first observe that construction of usefully concise encoding schemes for stable extensions is trivial: 
the problem of deciding, given $\HC({\mathcal X},{\mathcal A})$ and $S\subseteq{\mathcal X}$
whether $S\in\SEC(\HC)$ is polynomial-time solvable; therefore since $\HC({\mathcal X},{\mathcal A})$
already defines a terse encoding of $\SEC(\HC)$ with an appropriate decision algorithm
we have a concise $\SEC$ encoding scheme. This can be extended to give a usefully concise
scheme, by encoding $\HC$ as its $n^2$ element adjacency matrix, so that $w=\eta(H)$ is
decided in $|w|$ steps.

In contrast to the easy construction above, usefully concise encoding schemes for {\em preferred} extensions 
are `unlikely' to exist.
\begin{theorem}\label{thm:pe-bad}
If $\mbox{{\sc np}}\not=\mbox{{\sc co-np}}$ then usefully concise $\PEC$ encoding schemes do not exist.
\end{theorem}
\proof{
Suppose $\mbox{{\sc np}}\not=\mbox{{\sc co-np}}$ and that for the sake of contradiction, 
$\tuple{\eta,P,Q}$ is a usefully concise $\PEC$ encoding scheme. We show that $\tuple{\eta,P,Q}$
can be used as the basis of an {\sc np} decision method for {\sc pref-ext}.
Since this problem is {\sc co-np}--complete
it follows that the existence of such a decision method would imply $\mbox{{\sc np}}=\mbox{{\sc co-np}}$.
Since $\tuple{\eta,P,Q}$ is terse there is some constant $k$
such that $|\eta(\HC)|\leq n^k$ for any $n$-argument system $\HC$. 
Our {\sc np} algorithm is as follows:  given an instance $\tuple{\HC({\mathcal X},{\mathcal A}),S}$ of 
{\sc pref-ext} non-deterministically choose a sequence $\beta(\HC)$ of (at most) $|{\mathcal X}|^k$ bits. 
Then simulate $Q$ on input $\tuple{\HC,\beta(\HC)}$. If $\beta(\HC)=\eta(\HC)$ $Q$ will have an
accepting computation of polynomial length (since $Q$ is effective). Finally,
the program, $P$, is run with input $\tuple{\beta(\HC),S}$. Since $|\beta(\HC)|\leq |{\mathcal X}|^k$
and $P$ is a deterministic polynomial time computation, the (non-deterministic) algorithm
runs in time polynomial in ${\mathcal X}$. To see that the algorithm accepts instances
for which $S\in\PEC(\HC)$ it suffices to observe that {\em if} $S\in\PEC(\HC)$ then there is {\em some}
choice of $\beta(\HC)$ that will correspond to $\eta(\HC)$, be accepted by $Q$ and on which the
extension tractable algorithm $P$ will accept $\tuple{\beta(\HC),S}$.}

It should be noted that the argument used in the proof requires the assumption that $\tuple{\eta,P,Q}$
is effective. The reason being that if $P$ were invoked directly on the word $\beta(\HC)$ then
{\em every} instance would be accepted: given a subset $S$ there is certainly some argument
system, ${\mathcal G}$ for which $\PEC({\mathcal G})=\set{S}$ -- the system with $|S|$ isolated
arguments -- and if $\beta(\HC)=\eta({\mathcal G})$ then the instance is accepted regardless of
whether $S\in\PEC(\HC)$. 

Theorem~\ref{thm:pe-bad} indicates that even if one is prepared to invest considerable
computational effort in {\em constructing} an encoding $\eta(\HC)$, such effort will
not aid in testing $S\in\PEC(\HC)$ if the encoding form is terse {\em and} effective.

The problem, {\sc pref-ext-inf} in allowing knowledge regarding the set of credulously accepted arguments
to be given for free, can be seen as defining an alternative {\em terse} encoding scheme.
We note that this scheme is {\em not} effective (in our usage) (assuming $\mbox{{\sc np}}\not=\mbox{{\sc co-np}}$)
since in the encoding $\tuple{\HC,\alpha}$, should any bit of $\alpha$ be $\bot$
indicating the associated argument is {\em not} credulously accepted, we cannot
test $w=\eta(\HC)$ using an {\sc np} computation.

Thus, since determining the set of credulously accepted arguments in a system, may at worst
involve similar computational expenditure to that of enumerating preferred extensions,
the complexity of the problem {\sc pref-ext-inf} is of some interest: the implied encoding
scheme is not one which is within the scope of Theorem~\ref{thm:pe-bad}. Our next result
shows that {\sc pref-ext-inf} is no easier than {\sc pref-ext}.

\begin{theorem}\label{thm:inf-is-bad}
{\sc pref-ext-inf} is {\sc co-np}--complete.
\end{theorem}
\proof{Membership in {\sc co-np} is immediate from the fact that $\mbox{{\sc pref-ext}}\in\mbox{{\sc co-np}}$.
To show that {\sc pref-ext-inf} is {\sc co-np}--hard, we give a reduction from the problem
of deciding is a propositional formula in 3--{\sc cnf} is {\em unsatisfiable}: {\sc 3-unsat}.
Let
\[
\Phi(X_n) = \AND_{i=1}^{m} C_i\ \ =\ \ \AND_{i=1}^{m} (\ y_{i,1}\Or y_{i,2}\Or y_{i,3}\ )
\]
be an instance of {\sc 3-unsat}, so that each $y_{i,j}$ is a literal from 
$\set{x_1,\ldots,x_n,\neg x_1,\ldots,\neg x_n}$. Let $\set{x_{n+1},x_{n+2}}$ be two new propositional
variables (i.e. not belonging to $X_n$) and consider the {\sc cnf} formula $\Psi (X_n,x_{n+1},x_{n+2})$
defined as,
\[
\AND_{i=1}^{m} ( C_i\Or\neg x_{n+1}\Or x_{n+2})\And (C_i\Or x_{n+1}\Or\neg x_{n+2})
\]
The following properties of $\Psi(X_n,x_{n+1},x_{n+2})$ are easily verified.
\begin{enumerate}
\item[a)]
For each variable $y\in\set{X_n,x_{n+1},x_{n+2}}$ there is {\em some} satisfying instantiation for $\Psi$
under which $y=\top$.
\item[b)]
For each variable $y\in\set{X_n,x_{n+1},x_{n+2}}$ there is {\em some} satisfying instantiation for $\Psi$
under which $y=\bot$.
\item[c)]
There is a satisfying instantiation for $\Psi$ under which $x_{n+1}=\top$ and $x_{n+2}=\bot$ if and only if
$\Phi(X_n)$ is satisfiable.
\end{enumerate}
We use $\Psi(X_n,x_{n+1},x_{n+2})$ to build an argument system $\HC_{\Psi} ({\mathcal V},{\mathcal A})$.
The argument set ${\mathcal V}$ contains $2(m+n+3)$ arguments labelled
\[
\begin{array}{rcl}
V=\set{\Psi,\chi}&\cup&\set{ x_i , \bar{x}_i : 1\leq i \leq n+2}\\
&\cup&\set{ C_{j}^{(1)}, C_{j}^{(2)}\ :\ 1\leq j \leq m}
\end{array}
\]
The attack relationship, ${\mathcal A}$, comprises
\begin{enumerate}
\item
$\set{\tuple{x_i,\bar{x}_i},\tuple{\bar{x}_i,x_i}\ :\ 1\leq i\leq n+2}$
\item
$\set{\tuple{\chi,\bar{x}_i}, \tuple{\chi,x_i}\ :\ 1\leq i\leq n}$
\item
$\set{\tuple{y_{i,j},C_{i}^{(k)}}\ :\ 1\leq k\leq 2,\ 1\leq i\leq m,\ y_{i,j}\in C_i}$
\item
$\set{\tuple{C_{i}^{(1)},\Psi},\tuple{C_{i}^{(2)},\Psi}\ :\ 1\leq i\leq m}$
\item
$\set{\tuple{\bar{x}_{n+1},C_{i}^{(1)}},\tuple{x_{n+2},C_{i}^{(1)}}\ :\ 1\leq i\leq m}$
\item
$\set{\tuple{x_{n+1},C_{i}^{(2)}},\tuple{\bar{x}_{n+2},C_{i}^{(2)}}\ :\ 1\leq i\leq m}$
\item
$\set{\tuple{\Psi,\chi}}$.
\end{enumerate}
In the system $\HC_{\Psi} ({\mathcal V},{\mathcal A})$ all {\em except} the arguments
\[
\set{\chi}\cup\set{C_{j}^{(1)},C_{j}^{(2)}\ :\ 1\leq j\leq m}
\]
are credulously accepted. To see this first observe that a satisfying instantiation $\alpha$ of 
$\Psi(X_n,x_{n+1},x_{n+2})$ induces a preferred extension of $\HC_{\Psi}$ containing
$\Psi$ together with the $n+2$ arguments corresponding to literals evaluating to $\top$
under $\alpha$. From properties (a) and (b) above we deduce that $\Psi$ is credulously
accepted and each of the arguments $x_i$, $\bar{x}_i$ for $1\leq i\leq n+2$. In order for an argument
$C_{i}^{(k)}$ to be credulously accepted, an admissible set containing it, would have to
contain $\bar{y}_{i,j}$ for each $y_{i,j}\in C_i$. These arguments, however, are attacked by $\chi$:
$\Psi$ (the only attacker of $\chi$) cannot be included in an admissible set containing $C_{i}^{(k)}$.
Similarly $\chi$ cannot be credulously accepted, since the only attackers of its attacker - $\Psi$ -
are $C_{i}^{(k)}$. 

The instance of {\sc pref-ext-inf} constructed from $\Phi(X_n)$ is
$\tuple{\HC_{\Psi},\set{x_{n+1},\bar{x}_{n+2}},\alpha_{\mbox{{\sc ca}}}}$ where 
$\alpha_{\mbox{{\sc ca}}}$ is the $2(m+n+3)$ tuple indicating the credulously accepted arguments
in $\HC_\Psi$ as described in the preceding paragraph.

We claim that $\set{x_{n+1},\bar{x}_{n+2}}\in\PEC(\HC_\Psi)$ if and only if $\Phi(X_n)$ is unsatisfiable.
First suppose that $\set{x_{n+1},\bar{x}_{n+2}}\in\PEC(\HC_\Psi)$. Assume, for the sake of
contradiction, that $\Phi(X_n)$ is satisfiable and let $\alpha$ be a satisfying instantiation
of $X_n$ for $\Phi$. Consider the subset, $S_\alpha$ of ${\mathcal V}$ given by,
\[
\set{x_{n+1},\bar{x}_{n+2}}\cup\set{\Psi}\cup\set{x_i\ :\ \alpha_i=\top}\cup\set{\bar{x}_i\ :\ \alpha_i=\bot}
\]
We claim that $S_{\alpha}$ is an admissible set (in fact $S_{\alpha}\in\PEC(\HC_\Psi)$). To see this
first observe that $S_{\alpha}$ is conflict-free and consider any argument $p\in{\mathcal V}$ that
attacks some argument of $S_{\alpha}$. If $p=\chi$, then $\Psi\in S_\alpha$ attacks $p$;
if $p=y\in\set{x_i : 1\leq i\leq n+2}$, then $\bar{y}\in S_{\alpha}$ attacks $p$; similarly
if $p=\bar{y}\in\set{\bar{x}_i : 1\leq i\leq n+2}$ then $y\in S_\alpha$ attacks $p$. If 
$p\in\set{C_{i}^{(2)} : 1\leq i\leq m}$ then $x_{n+1}$ attacks $p$. We are left with the
case $p\in\set{C_{i}^{(1)}}$. Since $\alpha$ satisfies $\Phi$ some literal $y_{i,j}$ of $C_i$
must take the value $\top$ under $\alpha$. Now we find an attack on $C_{i}^{(1)}$ with
the corresponding $x_i$ or $\bar{x}_i$ argument in $S_{\alpha}$. We deduce that
$\Phi(X_n)$ satisfiable would contradict the assumption $\set{x_{n+1},\bar{x}_{n+2}}\in\PEC(\HC_{\Psi})$.

On the other hand, suppose that $\Phi(X_n)$ is unsatisfiable. We show that 
$\set{x_{n+1},\bar{x}_{n+2}}\in\PEC(\HC_\Psi)$. 

Certainly $\set{x_{n+1},\bar{x}_{n+2}}$ is admissible.
Consider any $S\subset{\mathcal V}$ for which $\set{x_{n+1},\bar{x}_{n+2}}\subseteq S$ and with
$S$ admissible. Let $p\in S/\set{x_{n+1},\bar{x}_{n+2}}$. If $p=y\in\set{x_i,\bar{x}_i:1\leq i\leq n}$,
then in order to counter-attack the attack by $\chi$ on $y$, the argument $\Psi$ must be in $S$.
If $\Psi\in S$, then for each argument $C_{i}^{(1)}$, $S$ must contain some argument corresponding
to a literal $y_{i,j}$ of $C_i$. Since $S$ is assumed admissible, it follows that the set
of literals identified do not conflict. Choosing an instantiation of $X_n$ which makes
each of these literals take the value $\top$ will satisfy $\Phi$. This, however, contradicts the
assumption that $\Phi$ was unsatisfiable. We deduce that $S/\set{x_{n+1},\bar{x}_{n+2}}=\emptyset$
and therefore $\set{x_{n+1},\bar{x}_{n+2}}\in\PEC(\HC_{\Psi})$ as claimed.}

The following Corollary is easily obtained,
\begin{corollary}\label{corr:stab-ext-inf}
{\sc stab-ext-inf} is {\sc np}--complete.
\end{corollary}
\proof{Using the construction of the Theorem, $\set{x_{n+1},\bar{x}_{n+2}}$ can be developed to
a {\em stable} extension, if and only if $\Phi(X_n)$ is satisfiable.}

It is stressed that Theorem~\ref{thm:inf-is-bad} and its corollary are addressing {\em different}
decision problems from their counterparts {\sc pref-ext} and {\sc stab-ext}: for the latter
problems an instance comprises an argument system $\HC(\XC,\AC)$ and a subset $S$ of $\XC$; in the
problems {\sc pref-ext-inf} and {\sc stab-ext-inf} an instance additionally provides 
$|\XC|$ bits of information, $\alpha_\XC$, delineating which arguments of $\XC$ are credulously accepted.
One indication of the different nature of these problems can be seen by considering the case when
$S=\emptyset$: {\sc pref-ext} is {\sc co-np}--complete for this case, however {\sc pref-ext-inf}
has an easy polynomial time algorithm by simply checking if any bit of $\alpha_\XC$ is $\top$. It is of
interest to note, however, that the complexity of {\sc stab-ext-inf} when $S=\emptyset$ is less
clear: while we can deduce the absence of any stable extension in $\HC$ from $\alpha_\XC=\tuple{\bot}^{|\XC|}$,
we cannot deduce that one {\em does} exist if $\alpha_{\XC}$ is not of this form. We conjecture
that, in fact, {\sc stab-ext-inf} is {\sc np}--complete even for instances $\tuple{\HC,\emptyset,\alpha_\XC}$.
We note that if correct, this provides a rare, albeit arguably `unnatural', example
of a problem where a decision concerning
preferred extensions is `easier' than the corresponding decision regarding {\em stable} extensions.

\section{Conclusions}
The principal focus of this article has been in deriving negative results concerning
various mechanisms for reducing the complexity of deciding $S\in\PEC(\HC)$ though
`expensive' precomputation. Thus for a rather general notion of `useful encoding scheme'
it has been shown that such approaches are unlikely to succeed. There remains,
of course, the possibility that specific sub-classes of argument system are amenable
to concise encoding approaches.

Another direction for further work arises from the fact that
the requirement for $\tuple{\eta,P,Q}$ to be {\em effective} is rather strong and, as we have
seen in Theorem~\ref{thm:inf-is-bad} does not apply to what might be regarded as otherwise
`reasonable' approaches. One possible encoding approach concerning which Theorem~\ref{thm:pe-bad}
will not in general apply is the following.

Recall that $f_{\HC} (X_n)$ is the propositional logic {\em function} for which $f_{\HC}(\alpha_S)=\top$
if and only the subset $S$ of ${\mathcal X}$ indicated by $\alpha_S$ is in $\PEC(\HC)$.
Given $f_{\HC} (X_n)$ one might represent this using a suitable {\em propositional formula}.
Of course, there are (infinitely) many equivalent formulae in this regard. Suppose
for a propositional formula $\Phi (X_n)$ over the (binary operation) basis of $\set{\And,\Or,\neg}$ we define
the {\em length} of $\Phi(X_n)$ as its total number of occurrences of literals., denoting this $|\Phi(X_n)|$.
Now consider the following measures.
\[
L (\HC) =_{def}\ \min\set{|\Phi(X_n)|\ :\ \mbox{$\Phi(X_n)$ represents $f_{\HC} (X_n)$}}
\]
\[
L(n) =  \max_{\HC({\mathcal X}_n,{\mathcal A})}\ \set{L(\HC)}
\]
Informally, $L(n)$ is given by: for each different $n$-argument $\HC$ identify the
shortest formula representing $f_{\HC}$; $L(n)$ is then the maximum of these values.
The function $L(n)$ is well-defined (and computable, albeit by highly infeasible mechanisms).
It is certainly the case that encoding $\PEC(\HC)$ by a propositional formula $\Phi(X_n)$
is an {\em extension tractable} approach: to test $S\in\PEC(\HC)$ simply evaluate $\Phi(\alpha_S)$.
This is unlikely to be {\em effective}: given $w\in\set{0,1}^*$, even though (assuming some
standard encoding of propositional formulae, e.g. \cite[p.273]{Dunne88}) one could
determine whether $w$ encodes {\em some} $\Phi$, it is unlikely that one can test within
{\sc np} if this represents $f_{\HC}$. In summary we have {\em open} the possibility
that propositional formulae offer {\em concise} encodings of $\PEC(\HC)$ since such
are not ruled out by Theorem~\ref{thm:pe-bad}. 
\begin{problem}\label{problem:prop-form}
Do propositional formulae admit concise $\PEC$ encodings, i.e. is there any $k\in\Nat$ such that
$L(n)=O(n^k)$?
\end{problem}

There are two points worth considering concerning Problem~\ref{problem:prop-form}. 
First, the classic information-theoretic argument of Riordan and Shannon\cite{Shannon:1942},
(cf \cite[pp.273--274]{Dunne88}) does not help in proving superpolynomial lower bounds:
even if it is assumed that each distinct $n$ argument
system encodes a different preferred extension set, the lower bound on $L(n)$ implied by this is only
$n^2/\log n$. A second point concerns the class of
propositional functions being addressed: work of Lupanov\cite{Lupanov:1962} indicates how
formulae for propositional functions satisfying certain `inheritance'
properties can be constructed. This approach -- the so-called ``Principle of Local Coding''
-- allows `small' formulae to be built for suitable classes of functions provided that
specific small formulae used in the approach can also be built. An overview
of the mechanism is given in \cite[Chapter 3, pp.~136--8]{Dunne88}. If one considers
the class ${\mathcal G}=\cup_{n=1}^{\infty} \set{{\mathcal F}_n}$ in which ${\mathcal F}_n$ is the set of
$n$-variable propositional functions $f_{\HC}$ for systems $\HC$ of $n$ arguments, then it may
be possible to show that ${\mathcal G}$ has the required `inheritance' property and that this,
given suitable subsidiary formula constructions {\em might} lead to concise encoding schemes. It should
be noted, however, even if this route is possible, it is likely to be the case that
generating and verifying the correctness of resulting formulae may well be computationally
demanding (although such a process need only be performed {\em once} with respect to any given
$\HC$).

As a final open question we mention the following decision problem. As our starting point for
building an encoding of $\PEC(\HC)$ we have assumed that the argument system $\HC$ is provided
as the instance. One might `relax' this and assume that an arbitrary {\em subset} of $\wp(X_n)$ is given
and we wish to encode {\em only} those subsets that correspond to $\PEC(\HC)$ for {\em some} 
$\HC({\mathcal X},{\mathcal A})$. Thus, we have the following decision problem:
\newline
{\sc realisable}
\newline
{\bf Instance}: ${\mathcal S}=\set{S_1,S_2,\ldots,S_k}\subseteq\wp({\mathcal X}_n)$.
\newline
{\bf Question}: Does there exist an argument system $\HC({\mathcal X}_n,{\mathcal A})$ for which 
$\PEC(\HC)={\mathcal S}$?
\begin{problem}
Determine the complexity classification of {\sc realisable}.
\end{problem}

We conjecture that $\mbox{{\sc realisable}}\in\mbox{{\sc p}}$, which would follow by proving that
the following condition (which is easily shown to be necessary) is also sufficient for
${\mathcal S}$ to be realisable.
\[
\begin{array}{l}
\forall T\subseteq{\mathcal X}_n\forall\set{x,y}\subseteq T \\
(\exists S_i\in{\mathcal S}\ S_i\subset T)\Rightarrow
(\forall S_i\in{\mathcal S}\ \set{x,y}\not\subseteq S_i)
\end{array}
\]
This condition can be tested in polynomial-time simply by restricting $T$ to range over those
supersets of $S_i\in{\mathcal S}$ formed by adding a single new argument.
\bibliographystyle{plain}

\end{document}